\def\year{2021}\relax
\documentclass[letterpaper]{article} 
\usepackage{aaai21}  
\usepackage{times}  
\usepackage{helvet} 
\usepackage{courier}  
\usepackage[hyphens]{url}  
\usepackage{graphicx} 
\def\UrlFont{\rm}  
\usepackage{natbib}  
\usepackage{caption} 
\usepackage{soul}
\usepackage{siunitx}
\usepackage{booktabs}
\usepackage{multirow}
\usepackage{amssymb,amsthm,enumitem}
\usepackage{amsmath}
\usepackage[switch]{lineno}

\title{Detection and Prediction of Nutrient Deficiency Stress using Longitudinal Aerial Imagery}
\author{
Saba Dadsetan\textsuperscript{\rm 1,2}\thanks{work done while at Intelinair, Inc.}, 
Gisele Rose\textsuperscript{\rm 1},
Naira Hovakimyan\textsuperscript{\rm 1,3},
Jennifer Hobbs\textsuperscript{\rm 1}\\
}
\affiliations {
\textsuperscript{\rm 1}Intelinair, Inc., Champaign, IL 61820\\
\textsuperscript{\rm 2}University of Pittsburgh, School of Computing and Information, Pittsburgh, PA 15213 \\
\textsuperscript{\rm 3}University of Illinois at Urbana Champaign, Department of Mechanical Science and Engineering, Urbana, IL 61801\\
sabadad@cs.pitt.edu, gisele@intelinair.com, naira@intelinair.com, jennifer@intelinair.com
}

\begin{document}

\maketitle

\begin{abstract}
Early, precise detection of nutrient deficiency stress (NDS) has key economic as well as environmental impact; precision application of chemicals in place of blanket application reduces operational costs for the growers while reducing the amount of chemicals which may enter the environment unnecessarily.  
Furthermore, earlier treatment reduces the amount of yield loss and therefore boosts crop production during a given season.
With this in mind, we collect sequences of high-resolution aerial imagery and construct semantic segmentation models to detect and predict NDS across the field; our work sits at the intersection of agriculture, remote sensing, and deep learning.
First, we establish a baseline for full-field detection of NDS and quantify the impact of pretraining, backbone architecture, input representation, and sampling strategy. 
We then quantify the amount of information available at different points in the season by building a single-timestamp model based on a U-Net. 
Next, we construct our proposed spatiotemporal architecture, which combines a U-Net with a convolutional LSTM to accurately detect regions of the field showing NDS; this approach has an impressive IOU score of 0.53.
Finally, we show that this architecture can be trained to \textit{predict} regions of the field which are expected to show NDS in a later flight- potentially more than three weeks in the future- maintaining an IOU score of 0.47-0.51 depending on how far in advance the prediction is made.
We will also release a dataset which we believe will benefit the computer vision, remote sensing, and agriculture fields.
This work contributes to the recent developments in deep learning for remote sensing and agriculture while addressing a key social challenge with implications for economics and sustainability. 
\end{abstract}

\section{Introduction}
Precision agriculture is a key rising area of interest for the application of deep learning approaches.
Computer vision approaches and applications in agriculture simultaneously address key social needs while furthering our understanding of the machine learning field by addressing unique theoretical and computational challenges.

Precision agriculture and sustainable practices are central to addressing challenges around economic hardship, food and clean water scarcity, and climate change~\cite{rolnick2019_bengioClimate}.
The Food and Agriculture Organization (FAO) of the United Nations estimates that to feed the world's ever-growing population 50\% more food needs to be produced by 2050~\cite{fao2017_future}.
This demand for an increase in production comes amidst challenges brought about by environmental changes as agriculture is both impacted by and contributes to climate change and other environmental issues.

\begin{figure}
    \centering
    \includegraphics[width=.5\linewidth, trim={0 0cm 0 0cm}]{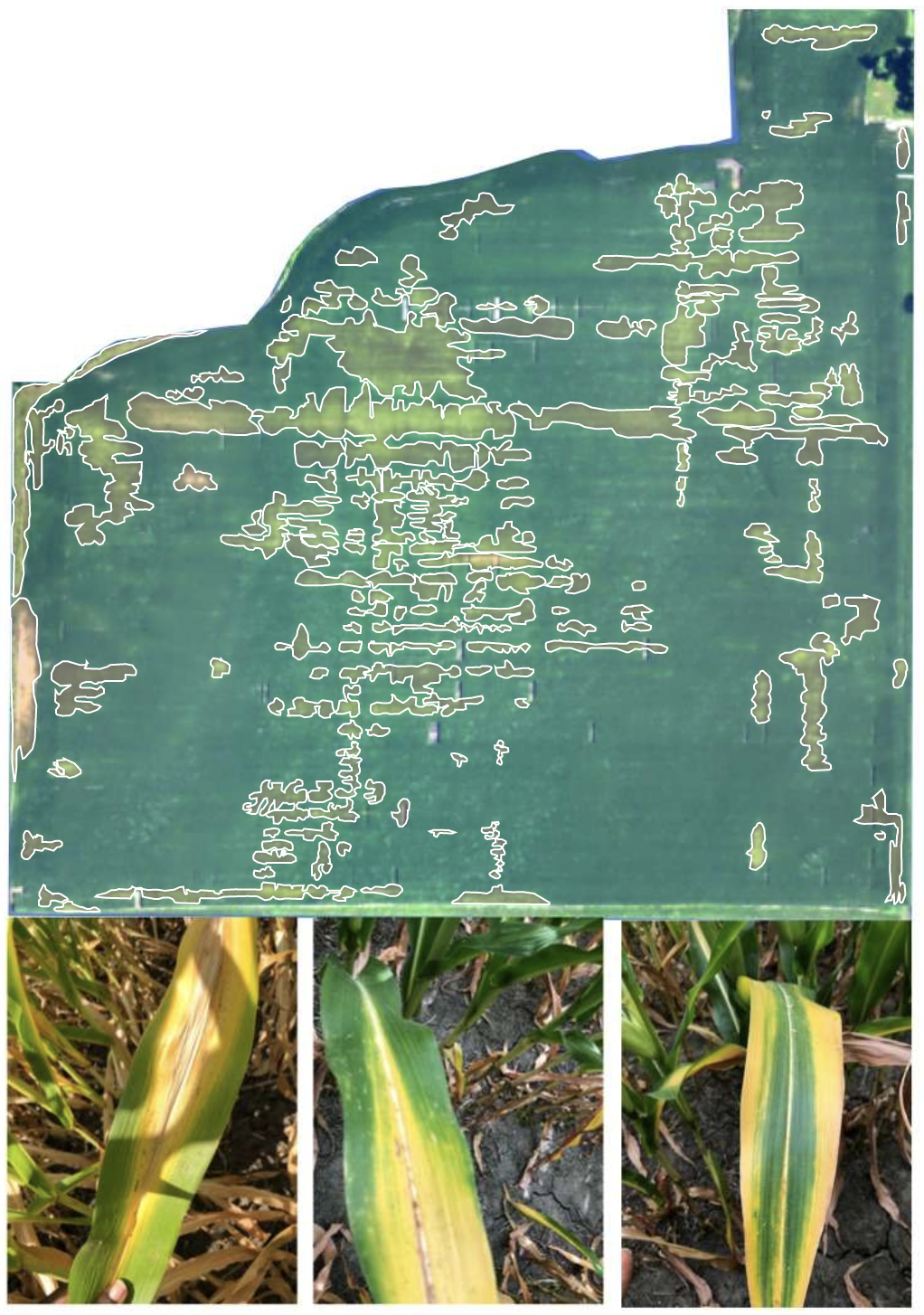}
    \caption{Nutrient Deficiency Stress appear brighter green in this aerial image which have been annotated with polygons(top). By alerting farmers to regions of the field experiencing NDS, further ground-level inspection can reveal what type of nutrient deficiency exists so the best management decisions can be made. Ground-level images of interveinal (bottom left), nitrogen (bottom middle), and potassium (bottom right) stress.}
\label{fig:nd}
\end{figure}

Central to these challenges is the task of identifying Nutrient Deficiency Stress (NDS) (Figure~\ref{fig:nd}).
Once significant NDS has set in, the ability for the crop to recover and produce full yield is minimal.  
As a result, early detection of regions experiencing NDS is paramount to ensuring an optimal yield; better still is the ability to predict which regions may begin showing stress in the near future so they may be preemptively treated. 
On the other hand, overuse and blanket applications of fertilizer and other chemical nutrients can cause harm to the plants and additionally, runoff into the water table or other bodies of water, harming the environment.
Therefore, early, precise and accurate identification of these regions is crucial for addressing both economic and environmental issues.
We choose to focus on this task for the present work although the methods could easily be applied to other agronomic patterns such as weeds or low emergence.  

\begin{figure*}
    \centering
    \includegraphics[width=1.0\linewidth, trim={0 0cm 0 0cm}]{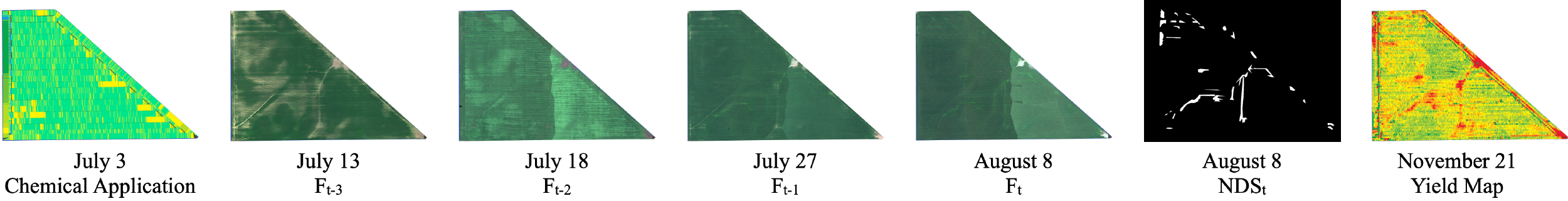}
    \caption{Temporal view of a field in this study.  Corn was planted on June 2 and a uniform application of fertilizer and weed repellent applied on July 3 (far left).  As the crop emerges and develops, under-performing areas due to nutrient deficiency stress (NDS) and other causes become visible during mid-season.  Unless treated early, these nutrient deficient areas eventually under produce at harvest time(red, far right). Our model can be used to both detect the presence of NDS in a given flight $F_t$ and predicts NDS in those areas from earlier flights $F_{t-3:t-1}$.}
\label{fig:intro}
\end{figure*}

Figure~\ref{fig:intro} shows the impact of NDS on a field's production and thus a farmer's yield.
On June 2, a farmer sowed over 4.5million corn seeds across his 152 acre farm in the corn belt.
Roughly a month later on July 3, he applied a combination of weed killer and fertilizer uniformly across the entire field (far left).  
By collecting high-resolution imagery (10cm/pixel) over the entire season, we are able to observe the changes in the field as it develops (middle plots).
We see that many of the areas which experienced NDS during the season correspond to regions of lower yield (red) at harvest time in late November (far right); these red areas produced $>$195 bushels/acre (orange: 195-230 bu/ac) whereas the best areas (dark green) produced over 275 bushels/acre.  
Had the grower applied additional nutrient mid-season, some of this yield could have been recovered.  
However, as these areas are only a small portion of the field, blanket application would be costly as wasteful.
Through targeted applications based on our models, the grower could instead apply fertilizers in a precise manner, minimizing cost, maximizing yield, and minimizing the runoff of chemicals into the water system.

To bring targeted intelligence to the grower in an actionable time frame, we collect high-resolution aerial imagery multiple times across the season.
We first baseline our approach by building a simple U-Net-style segmentation model to detect areas of NDS. 
We then improve on this approach by leveraging the sequence of flights to create a spatiotemporal detection model.
Finally, we show this same architecture can be used to predict areas of NDS in subsequent flights.


\section{Related Work}

\subsection{Aerial Imagery, Remote Sensing, and Agriculture}
\label{rw:imagery}
Aerial Imagery and remote sensing techniques have been commonplace in the agriculture space for many years \cite{mulla2013_remoteSensingAg,maes2019_perspectives,clevers1986_remoteField,idso1977_remoteYield}.  
Applications are widespread including high-throughput phenotyping~\cite{araus2014field_phenotyping}, biomass prediction~\cite{biomasspred}, irrigation management~\cite{bastiaanssen2000remote_irrigation}, weed detection~\cite{thorp2004review_weed}, disease and pest detection~\cite{zhang2019monitoring_disease}, nitrogen status and usage~\cite{bausch1996remote_nitrogen, bausch2001innovative_nitrogen}, and many others.

In applications related to agriculture, vegetative indices like the Normalized Difference Vegetative Index (NDVI)~\cite{sharma2015active_ndvi} have been central to traditional computer vision based analyses~\cite{xue2017_indices,huete2002_indices2}.
Another key index, Green Normalized Difference Vegetative Index (GNDVI)~\cite{gitelson1996use_gndvi} is more strongly correlated with the concentration of chlorophyll and therefore to the rate of photosynthesis, making it a potential indicators of stress.
The Normalized Difference Water Index (NDWI)~\cite{mcfeeters1996use_ndwi} is related to the water content in bodies, such as plants, and therefore can be informative about water-related stress.

As with many algorithms based on hand-crafted features, algorithms based on vegetative indices suffer from appearance changes due to variable lighting; in the imagery of Figure~\ref{fig:intro} seamlines are present, resulting in different broad appearances on the two sides of the image.
This can cause challenges when relying on these indices unless other corrections and adjustments are made~\cite{rodriguez2006detection_ndvi, noh2005dynamic_ndvi, mamaghani2018initial}.
We explore the use of indices for this deep learning task in \textit{Experiments and Results: Vegetative Indices}.

\subsection{Nutrient Deficiency Identification}
Both traditional computer vision and deep learning methods have been used to detect and classify types of nutrient deficiency stress.
Many of these results focus on identifying the type of deficiency from close-up images of the plant~\cite{sartin2014image_nd, sethy2020nitrogen_leaf}.
Early work on aerial imagery focused on identifying signatures in hyperspectral imagery and shortwave radiation correlated with the presence of NDS~\cite{goel2003potential_ndrm, blackmer1995remote_ndrm}.
Most of these cite changes in the 380-720nm or 720-1500nm range as strong indicators of stress due to changes in chlorophyll activity~\cite{mee2017detecting_ndsReview}.
To the best of our knowledge, no work has been done to \textit{forecast} NDS directly from aerial imagery.

\subsection{Deep Learning in Agriculture}
The adoption of deep learning methods for agricultural applications has accelerated in recent years~\cite{kamilaris2018_DLInAgSurvey,liakos2018_mlagReview}.
These results can be largely split into those focused on ``standard'' imagery and those focused on aerial imagery from satellite, drone, or aircraft.
Applications include disease and pest identification~\cite{mohanty2016_dlDisease,wiesner2019_mmDisease,boulent19_diseasecnn}, crop identification~\cite{cvpr_cropType}, crop counting~\cite{malambo2019_sorghumcounting,li2017_palmoilcounting}, weed detection~\cite{sa2018_weedmap,bah2018deepweeds,sa2017weednet}, yield forecasting~\cite{barbosa2020modeling,nevavuori2019_cropYield},
, and parcel segmentation~\cite{aung2020farmland} as a few examples. 

Specifically relevant to this work is~\cite{chiu2020agriculture} which used deep learning-based segmentation techniques to semantically segment the field into different patterns, including NDS, from high-resolution aerial imagery.
Their approach used a DeepLabV3+~\cite{deeplabv3plus2018} model for their segmentation task.
We similarly use an encoder-decoder structure, but focus on a U-Net~\cite{unet} framework because of its success in other agricultural applications and computational efficiency~\cite{unet_counting,chiu20201st}.

\subsection{Spatiotemporal Modeling}
Encoder-Decoder models like U-Net are commonplace in modern semantic segmentation tasks because of their performance, speed, and flexibility.  
More recently it has become common to use different backbones such as EfficientNet~\cite{tan2019efficientnet} within the U-Net framework.
To address the temporal nature of data, Long Short Term Memories (LSTMs) ~\cite{hochreiter1997long_lstm} are frequently employed in a variety of deep learning-based sequence tasks including handwriting recognition, language translation, and action recognition, as a few examples ~\cite{graves2008novel_lstm1, wu2016google_lstm2}.

Spatiotemporal modeling in the remote sensing domain is an active area of research~\cite{zhu2017deepReview}.  
Methods for handling the temporal element are highly varies and include: spatiotemporal U-Net~\cite{lin2020farm}, histogram-based input representations~\cite{you2017deepyield}, 3D convolutions~\cite{ji20183d_3dconv}, and many others.  
Our proposed method most closely resembles the Fully Convolutional Network(FCN)-LSTM network of ~\cite{teimouri2019novel_fcn-lstm}, and was similarly chosen for its efficiency and performance.

\section{Methods}
\subsection{Data Collection}
Much of the work done in remote sensing for agriculture uses either low-resolution (10m/pixel) satellite imagery or very high-resolution ($<$5cm/pixel) imagery obtained from drones.
While low-resolution satellite imagery provides a good overview of the field and scales across large areas, it does not provide enough resolution to precisely identify areas of the field which may exhibit stress.
Conversely, while the imagery from drones is much higher, it is impractical to use to gather the desired data at such a low-resolution across millions of acres in a region.

Therefore to collect our data, fixed-wing aircraft were flown across corn and soybean fields in Illinois, Indiana, and Iowa, capturing imagery up to 13 times across the 2019 growing season (April to October).
RGB and near-infrared (NIR) images were simultaneously captured at a resolution of 10cm/pixel using a Wide Area Multi-Spectral System (WAMS).

Mosaicking using ground-control points is performed to create a single large image per farm; depending on the farm size, this results in an image on average 15k-pixels$\times$15k-pixels in dimension.
Orthorectification is performed using the RGBN image and a digital elevation model (DEM) of the field to produce a plainmetrically correct image~\citep{orthorect}. Geo-information is tagged to every image, but not used in this work nor a part of the data release to protect privacy. 

\subsection{Annotation and Dataset Construction}
Images from 670 farm parcels were annotated for regions of nutrient deficiency stress by human experts; quality assurance (QA) of the annotations was conducted after.
We focus only on mid-season flights, 6-10, when NDS is potentially present. 
386 of the 670 flights contain at least 3 flights during this period and at least one flight demonstrating NDS; to conduct a fair comparison between single flight analysis and multiple flights analysis, we focus only on these 386 fields. 
The last annotated flight from each of the 386 fields becomes the target of subsequent analysis.
Those 386 flights contained 10052 regions of NDS in total; not every flight showed signs of NDS while many had multiple regions.
NDS is a relatively rare pattern spatially, resulting in an imbalanced dataset; on fields containing any NDS, an average 21\% of those pixels contained NDS.

As the original images contain over 225 million pixels on average, we reduced the size by converting to 300dpi- high enough to preserve the pixel information and low enough to fit into our memory during training. 
This results in images roughly 1000-to-2000 pixels by 1000-to-2000 pixels with 1m/pixel resolution, still far higher than most satellite imagery.

As a part of this work we are releasing these three flights from the 386 farm parcels as well as the ground-truth mask for that final flight\footnote{This dataset will be released on the Registry of Open Data on AWS under "Longitudinal Nutrient Deficiency".  Supplementary material available at \url{https://arxiv.org/abs/2012.09654}.}.

\subsubsection{Data Augmentations:}
During training we perform data augmentation such as vertical/horizontal flipping, shifting, rotation, and padding using the Albumentations package~\cite{info11020125_alb}. 
We do not perform any color-based augmentation because these images are narrow-band and therefore standard augmentation techniques do not produce the same results as in standard imagery.
Note that in experiments where we use temporal data, we perform the same augmentation on all the images in the sequence.  

\subsection{Models}
\textbf{Single Timestep:}
To perform either NDS detection or prediction for only one time-point we use a U-Net structure with a VGG16 or EfficientNet backbone. 
Given the single input data point $I$, we have $G$ as a ground truth mask and $P$ as a predicted mask. 
The size of input $I$ is w$\times$h$\times$c where c indicates the number of channels. 
Both the ground truth mask $G$ and predicted mask $P$ show the binary segmentation of NDS areas therefore they have a size of w$\times$h$\times$1.
We calculate the combined Focal loss~\cite{focal_loss} $+$ Dice loss~\cite{dice} for our final loss.

\begin{figure*}
    \centering
    \includegraphics[width=.65\linewidth, trim={0 0cm 0 0cm}]{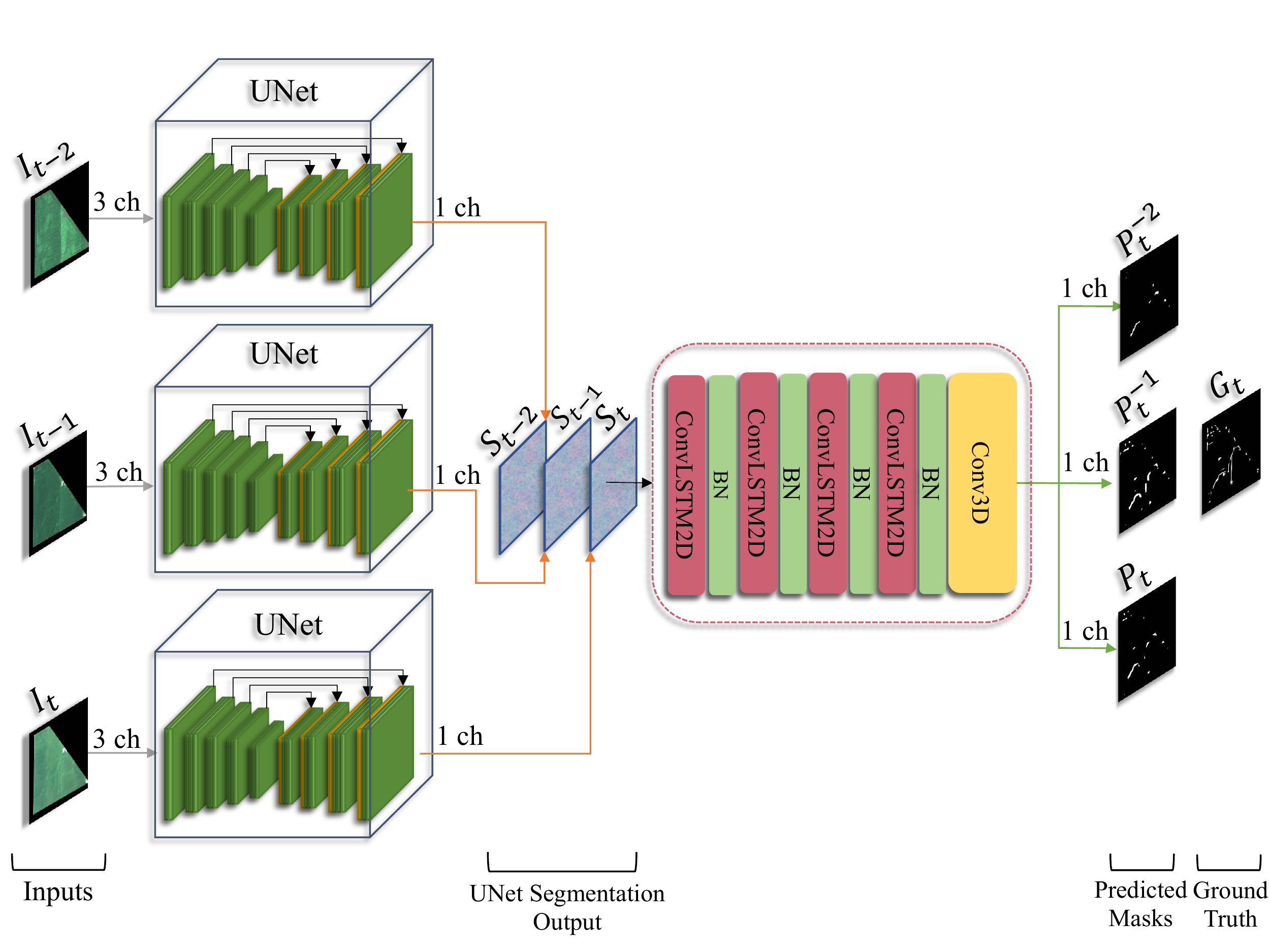}
    \caption{Our proposed model architecture for longitudinal detection of NDS. RGB images at three timesteps, $I_{t}, I_{t-1}, I_{t-2}$, are each passed through a U-Net to generate single channel outputs $S_t, S_{t-1}, S_{t-2}$. These outputs are stacked and then passed through a convolutional-LSTM layers which generates prediction $P_t, P_{t}^{-1}, P_{t}^{-2}$ of NDS for flight $t$ called $G_t$.}   
    
\label{fig:post_lstm}
\end{figure*}

\textbf{Multiple Timesteps:}
A key aspect of the collected data is its sequential nature which captures the evolution of the field over time. 
To incorporate the information from sequential flights, we construct the model seen in Figure \ref{fig:post_lstm}. This model include 3 parallel U-Nets with EfficientNet backbones followed by a sequence of 2D convolutional-LSTM and Batch-normalization(BN) layers and a 3D Convolution layer to generate the final output. 
Given 3 consecutive flight images $I_{t}$, $I_{t-1}$ and $I_{t-2}$ and the final ground truth mask $G_{t}$ belonging to time step $t$, each U-Net produces its respective binary mask $S_{t}$, $S_{t-1}$ and $S_{t-2}$. 
These masks are stacked and passed through a convolutional-LSTM (many-to-many) layers. Finally the last 3D convolution layer generate the semantic segmentation mask using a sigmoid function, identifying NDS for each flight denoted by $P_{t}$, $P_{t}^{-1}$ and $P_{t}^{-2}$.
To force the output of each U-Net to resemble the final mask, not just to provide information to the next flight via the LSTM, we calculate the combined Focal $+$ Dice loss for each of these predictions $P_t, P_{t}^{-1}, P_{t}^{-2}$. 
The total loss is used defined as:
\begin{align*}
    \mathcal{L}_{Total} &= \frac{1}{3} \sum_{i=0}^{2}Loss_{t-i}^{Focal} + Loss_{t-i}^{Dice}
\end{align*}

\subsubsection{Training: }
All experiments are conducted with a batch size of 2 for 200 epochs; all models reach their convergence point before the final epoch and the best model is chosen based on minimum validation loss.
We use Adam optimization with an initial learning rate of 1e-4 and decayed it by a factor of 10 if no improvement to the validation loss was seen for a period of 10 epochs. We use Intersection Over Union (IOU) score and F1 score as two evaluation metrics to compare models.
All the models are implemented using Keras (version 2.2.4) and Tensorflow (version 1.15) and we run them on 4 NVIDIA Tesla V100 GPUs with 64GiB memory in total.

\section{Experiments and Results}
Our experiments are as follows: first, we show different approaches to find the best model for single time step detection using latest flight and its NDS ground truth mask. In this evaluation, four scenarios have been considered including data augmentation, U-Net backbone, ImageNet pretraining and different vegetation indexes. In the next step, we show the results of our proposed model for multiple timesteps in both NDS detection and prediction along side with ablation studies results for comparison. 

\subsection{Single Timestep Baselines}
We explore the impact of resolution on model performance by either rescaling or cropping the images.
In the former, we rescale the full-field image to 512$\times$512 using bilinear interpolation.
Alternately, we crop the images to a fixed size of 512$\times$512, thereby maintaining the full resolution. 
Cropping is performed in the training pipeline in one of two ways: ``random'' or ``wise'' crop.
For random cropping, we select the 512$\times$512 patch from full-size image randomly, resulting in many patches with no nutrient deficiency.
In contrast, in our ``wise crop'' approach, we use a 512$\times$512 patch for training only if it contains some NDS masks.

ImageNet~\cite{deng2009imagenet} pretraining is common in computer vision applications because it can speed up training and may result in a better learned representation~\cite{kornblith2019better}.  
However, since the statistics of remote-sensing imagery in general and aerial agricultural imagery in particular are dramatically different than ImageNet~\cite{xie2015transfer}, we explicitly investigated its usefulness in pretraining in this domain.
Therefore we conduct all the experiments here with and without pre-trained ImageNet weights and compared their performance on the test dataset. 

For all the experiments we use U-Net framework and compare two different backbones: VGG16~\cite{simonyan2014_vgg} and EfficientNet-B5~\cite{tan2019efficientnet}.
We use data from the 386 flights randomly separated to: 231(60\%) train, 77(20\%) validation, and 78(\%20) test. We used only the RGB channels, ignoring the NIR channel.

\begin{table*}[hbt!]
\centering
\begin{tabular}{@{}clccc@{}}
\toprule
                                      &                            & \textbf{F1 Score} & \textbf{IOU Score} & \textbf{Loss (Focal + Dice)} \\ \midrule
\multirow{6}{*}{ImageNet Pretraining} & VGG16 Full Rescaled        & 0.33              & 0.25               & 0.76                         \\
                                      & VGG16 Random Crop          & 0.36              & 0.30               & 0.75                         \\
                                      & VGG16 Wise Crop            & 0.38              & 0.30               & 0.71                         \\
                                      & EfficientNet Full Rescaled & 0.39              & 0.26               & 0.65                         \\
                                      & EfficientNet Random Crop   & 0.40              & \textbf{0.33}      & 0.64                         \\
                                      & \textbf{EfficientNet Wise Crop}     & \textbf{0.43}     & \textbf{0.34}      & \textbf{0.58}                \\ \cmidrule(l){2-5} 
\multirow{6}{*}{No Pretraining}       & VGG16 Full Rescaled        & 0.28              & 0.19               & 0.80                         \\
                                      & VGG16 Random Crop          & 0.30              & 0.26               & 0.76                         \\
                                      & VGG16 Wise Crop            & 0.30              & 0.28               & 0.73                         \\
                                      & EfficientNet Full Rescaled & 0.31              & 0.21               & 0.74                         \\
                                      & EfficientNet Random Crop   & 0.33              & 0.29               & 0.69                         \\
                                      & \textbf{EfficientNet Wise Crop}     & \textbf{0.36}     & \textbf{0.29}      & \textbf{0.68}                \\ \bottomrule
\end{tabular}
\caption{Impact of backbone architecture, pretraining, and cropping on performance}
\label{table:baselines}
\end{table*}

As seen in Table~\ref{table:baselines}, the U-Net pretrained with ImageNet weights with an EfficientNet-B5 backbone using the ``wise cropping'' strategy produces the best results with an F1-score of 0.43, an IOU of 0.34, and a (Focal + Dice) loss of 0.58. 
While these results are slightly lower than those of \cite{chiu2020agriculture}, the two analyses are not directly comparable; the resolution and number of samples is lower in our analysis, we use only RGB instead of RGBN, and this is a single instead of multi-task approach.
The goal of our baselining is not to outperform other single-timestep models, but to quantify the performance on this particular dataset and establish an understanding of what architectural and sampling strategies might best guide our longitudinal analysis. 

Using pretrained ImageNet weights improves performance for each of the models.
Although the statistics of ImageNet and our dataset are quite different as noted before, the ImageNet weights nevertheless lead to an improvement in performance; therefore all subsequent models are conducted using pretrained weights.

Comparing the impact of backbones, our analysis shows that the EfficientNet backbone continually outperforms the VGG16 backbone; therefore we focus on EfficientNet as the backbone to our U-Net in our longitudinal analysis. 

Finally, these results show the importance of the wise crop sampling strategy over rescaling or random cropping.  
Regardless of the backbone and pretraining used, wise crop always lead to improved results; therefore we use this cropping strategy during our subsequent analyses.  
An example of the output from this best model is shown in the top row of Figure~\ref{fig:results}.

\subsection{Vegetative Indices}
\label{ex:indices}
Input representation is central to successful deep learning models.
Because vegetative indices play such a key role in traditional agricultural analysis and pattern detection, we next examine the impact and usefulness these indices have in a deep learning setting.
Definitions of theses indices are provided in the Supplementary Material.

Given the results of the previous analysis, we use 512$\times$512 images obtained with wise cropping, and a U-Net with an EfficientNet backbone, pretrained on ImageNet for all subsequent experiments. 
We compare four different input representations: RGB only (3 channel), NDVI (1 channel), RGB + NDVI (4 channel) with a 1D convolution applied at the first layer, and NDVI + GNDVI + NDWI (3 channel).
Results are shown in Table~\ref{table:ndvi}.

\begin{table*}[hbt!]
\centering
\begin{tabular}{@{}lccc@{}}
\toprule
                  & F1 Score & IOU Score & Loss (Focal +  Dice) \\ \midrule
Image (RGB)       & \textbf{0.43}     & \textbf{0.34}      & \textbf{0.58}                 \\
NDVI              & 0.25     & 0.15      & 0.85                 \\
Image(RGB)+NDVI & 0.36     & 0.29      & 0.69                 \\
Image(RGB)+GNDVI & 0.34     & 0.29      & 0.71                 \\
Image(RGB)+NDWI & 0.33     & 0.26      & 0.75                  \\
NDVI, GNDVI, NDWI & 0.32     & 0.24      & 0.73                 \\
\bottomrule
\end{tabular}
\caption{Impact of vegetative indices and image channels on performance}
\label{table:ndvi}
\end{table*}

These results show that using the raw RGB image as the input into the model far outperform any other input representation.
NDVI alone produce the worst results.  
This is perhaps not surprising because it contains information from only 2 channels (Red and NIR) compare to the 3 RGB channels so there is necessarily less information present.  
However, given its prevalence in the remote-sensing domain and the reliance on this metric in past approaches, quantifying just how much information is lost by using this single index is important.

Interestingly, using the combination of RGB and NDVI, which incorporates information from all four channels, with a 1D (channel-wise) convolution to create a new input representation for the U-Net, perform worse than RGB alone.  
We suspect this is due to losing too much information too quickly in this first 1D convolutional layer; that is, even by ``learning'' a new 1D representation as opposed to proscribing it through an index like NDVI, significant information is lost by quickly reducing the dimensionality.

Given the superior performance of the RGB representation, we focus on an RGB-only input for our longitudinal analysis.

\subsection{Detection of NDS using Longitudinal Data}
The best single timestep model from the earlier analysis is used to predict ${G_t}$ from a single image. However this time after initializing the network using ImageNet pretraining, we freeze the layers of encoder to decrease network's trainable parameters. 
This make single step analysis more comparable to the subsequent longitudinal study which are relatively larger and take more space from GPU's memory to fit.
Results are shown in Table~\ref{table:model_performance}.
As expected, the performance of the detection task ($P_t: I_t\rightarrow {G_t}$) is comparable to the previous results.
While information about ${G_t}$ is contained in $I_{t-1}$ and $I_{t-2}$ as seen in the prediction tasks $P_t^{-1}: I_{t-1}\rightarrow G_t$ and $P_t^{-2}: I_{t-2}\rightarrow G_{t}$, respectively, the performance decreases substantially to a point that would be unusable for any real-world application.
Note that the loss for these tasks corresponds to a loss for only one timestep whereas the other tasks in Table~\ref{table:model_performance} capture the average loss from three timesteps. 

Additionally, we stack all three flights to create a 9-channel image and again used the same U-Net framework.  
This model perform particularly poorly so we add a 1D Convolution after the input layer; this raise the performance to be only slightly better than the single image ($I_t$) model.

\begin{figure*}[ht]
    \centering
    \includegraphics[width=0.7\linewidth, trim={0 0cm 0 0cm}]{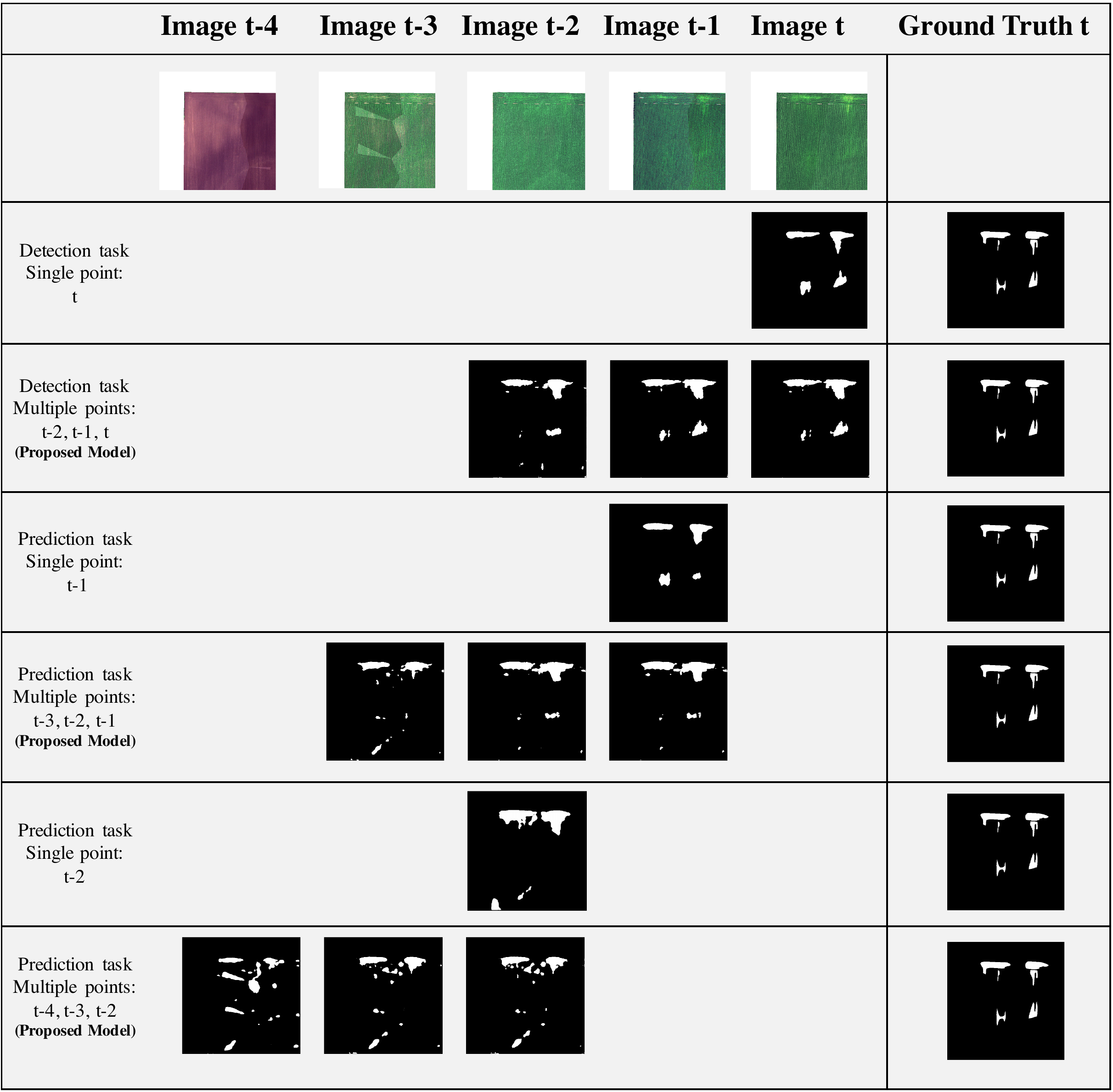}
    \caption{Results showing the predicted NDS mask $P_t,P_t^{-1},P_t^{-2}$ by using single flight model or sequential flights model (our proposed model).}   
\label{fig:results}
\end{figure*}

\begin{table*}[!h]
\centering
\begin{tabular}{@{}llccc@{}}
\toprule
                                          &                             & \textbf{F1 Score} & \textbf{IOU Score} & \textbf{Loss (Focal + Dice)} \\ \midrule
\multirow{10}{*}{\textbf{Detection Task: $I_{t},I_{t-1},I_{t-2} $}} & Single ($P_t:I_t\rightarrow G_t$)                     & 0.43              & 0.30               & 0.68                         \\ \cmidrule(lr){2-2}
                                          & 9-Channel                   & 0.23              & 0.15               & 0.90                         \\
                                          & 9-Channel + 1D Conv         & 0.48              & 0.34               & 0.62                         \\
                                          & Proposed- Unshared          & 0.58              & 0.53               & \textbf{0.85}                \\
                                          & \textbf{Proposed- Shared}            & \textbf{0.62}     & \textbf{0.57}      & \textbf{0.85}                \\
                                          & Only-LSTM                   & 0.48              & 0.43               & 0.90                         \\
                                          & Pre-LSTM + Concat           & 0.42              & 0.29               & 0.92                         \\
                                          & Pre-LSTM + Multi            & 0.42              & 0.29               & 0.89                         \\
                                          & Cascading-Model + Concat    & 0.43              & 0.30               & 0.87                         \\
                                          & Cascading-Model + Multi     & 0.38              & 0.25               & 0.91                         \\ \cmidrule(l){2-5} 
\multirow{3}{*}{\textbf{Prediction Task:  $I_{t-1},I_{t-2},I_{t-3} $}} & Single ($P_t^{-1}:I_{t-1}\rightarrow G_t$)                      & 0.38              & 0.26               & 0.73                         \\ \cmidrule(lr){2-2}
                                          & Proposed- Unshared          & 0.55              & 0.52               & 0.91                         \\
                                          & \textbf{Proposed- Shared}            & \textbf{0.57}     & \textbf{0.53}      & \textbf{0.90}                \\ \cmidrule(l){2-5} 
\multirow{3}{*}{\textbf{Prediction Task: $I_{t-2},I_{t-3},I_{t-4} $}} & Single   ($P_t^{-2}:I_{t-2}\rightarrow G_t$)                     & 0.27              & 0.18               & 0.84                         \\ \cmidrule(lr){2-2}
                                          & \textbf{Proposed- Unshared} & \textbf{0.48}     & \textbf{0.44}      & \textbf{0.93}                \\
                                          & Proposed- Shared            & 0.46              & 0.43               & \textbf{0.93}                \\ \bottomrule
\end{tabular}
\caption{Performance of our longitudinal models for detection and prediction}
\label{table:model_performance}
\end{table*}

To incorporate the information from sequential flights, we use our proposed model seen in Figure \ref{fig:post_lstm} and examine the impact of shared vs unshared weights of the U-Nets.
Additionally we compare this approach to three alternative approaches we call \textit{Only-LSTM}, \textit{Pre-LSTM} and \textit{Cascading-model}.
The Only-LSTM model contain only the LSTM part of our main proposed model.
The Pre-LSTM model takes a raw input images in sequence directly into the convolutional LSTM then multiplies (Hadamarad product) or concatenates the results with original inputs and passes them to 3 parallel U-Nets. 
The Cascading-model with concatenation takes an image $I_{t-2}$, passes it through a U-Net to get a predicted mask, and then combine the predicted probabilities $P_{t}^{-2}$ with the next image $I_{t-1}$ by concatenating it as a $4^{th}$ channel.  This is passed through a second U-Net to produce mask $P_{t}^{-1}$; this mask is concatenated with $I_{t}$, passes through another U-Net, and the final predicted mask $P_{t}$ is produced.  
Weights are updated such that the loss from $P_{t}$ is propagated through the entire network, those from $P_{t}^{-1}$ are propagated only through the first two U-Nets, and those from $P_{t}^{-2}$ are propagated only through the first U-Net.
The Cascading-model using multiplication module is the same as the previous except that the concatenations are replaced by the Hadamarad product.
The diagram of these models are shown in the Supplementary Material.

Our proposed approach outperform the alternative models across all metrics (Table~\ref{table:model_performance}).
The predicted masks at each timestep are shown in Figure~\ref{fig:results}.
The model with the shared weights slightly outperforms or matches the model with unshared weights on all metrics, but also has the advantage of being much smaller in size.
Unsurprisingly, incorporating all three flights significantly outperforms any of the single step models.  
While the improvement is not surprising, the amount of improvement is: the IOU is almost double that of the single-step model. 

One might expect that the current flight includes (almost) all necessary information for detection because it reflects the current status of the field.
However, as discussed in section \textit{Aerial Imagery, Remote Sensing, and Agriculture}, there are large changes to the global appearance of the field including natural development of the growing season as well as lighting and other noise effects.  
It is reasonable to believe that the sequence of images allows the model to better differentiate between features explaining changes due to the underlying NDS and these other sources of noise which impact larger regions of the field.

\subsection{Prediction of NDS using Longitudinal Data}
We next ask whether this architecture could be useful in \textit{predicting} NDS in later flights.
Using our same proposed architecture from the previous section, we train the model on images $I_{t-1}, I_{t-2}, I_{t-3}$ to predict the nutrient deficiency regions of $P_{t}$; we call this ``Prediction $t^{1:3}$''.
We go a step further and train another model on images $I_{t-2}, I_{t-3}, I_{t-4}$, again to predict the regions of $P_{t}$; we call this ``Prediction $t^{2:4}$''.
We also reference the single-step prediction tasks discussed earlier.

Even on these prediction tasks the spatiotemporal model does quite well (Table~\ref{table:model_performance}, Figure~\ref{fig:results}).
The loss for the the prediction task one flight out is $0.90$ with an IOU of $0.53$(shared) and the loss for two flights out is $0.93$ with an IOU of $0.43$(shared).
As in the detection task, using shared vs. unshared weights does not show a significant difference, so we prefer the shared weights because of the reduced model size.
Note that for this task of predicting two flights out, which can correspond to as much as 3 weeks into the future, the IOU of the predicted mask is better than even the best detection model made from a single image.
This suggests that incorporating the temporal element of this data and allowing the model to learn how the field evolves over time has tremendous value even beyond providing the stability we saw in the longitudinal detection analysis.

\section{Conclusion}
This paper addresses the important task of identifying nutrient deficiency stress from longitudinal aerial imagery.
The ability to not only detect, but \textit{predict} regions of NDS in a field has tremendous economic value to the farmers as well as environmental impact and sustainability efforts.
Our work shows that while a single image of the field is useful for detecting NDS, a sequence of images when used with an appropriate architecture provides significantly improved performance for both detection and prediction.
While the present work has focused only on nutrient deficiency stress, we believe this framework will be useful for detecting and predicting other patterns of interest such as weeds, drydown, water, and others.
Importantly, these models can be easily deployed at scale to a typical data pipeline for aerial agricultural imagery for maximum impact. 
They can easily be optimized for target hardware using frameworks which further improve inference speed and therefore reduce inference costs.

As this dataset is at a much higher resolution than most publicly available remote sensing imagery, we believe this will open the door to interesting future research.
We only began to explore the usefulness of the NIR channel through our examination of vegetative indices; this and other studies around alternate or learned vegetative indices is the focus of ongoing work.
While we explored a number of architectural variations, there is significant work being done on spatiotemporal data and sequential remote sensing imagery using completely different paradigms which this dataset will further enable. 
Our hope is that multiple research communities find this dataset useful in advancing both computer vision and sustainable agriculture.

\bibliography{refs}
\end{document}


\maketitle

\section{Definition of Vegetative Indices}
\begin{align*}
\begin{split}
 NDVI &= \frac{NIR-Red}{NIR+Red}\\
 GNDVI &=\frac{NIR-Green}{NIR+Green}\\
 NDWI &= \frac{Green-NIR}{NIR+Green}
 \end{split}
 \end{align*}
 where Red, Green, and NIR correspond to the reflectence values in those channels.





\section{Supplemental Tables}
\begin{table}[H]
\centering
\caption{Performance of our \textbf{proposed model} architecture for \textbf{detection} at each timestep output by the LSTM}
\label{table:proposed_timestep}
\begin{tabular}{llccc}
\hline
                                      & & \textbf{F1 Score} & \textbf{IOU Score} & \textbf{Loss (Focal + Dice)} \\ 
                                      \hline
\multirow{ 3}{*}{\textbf{Proposed- Unshared}} & Step $t$ & 0.58   & 0.53  & 0.85\\
                            & Step $t-1$          & 0.58   & 0.53   & 0.85  \\
                            & Step $t-2$           &  0.53  & 0.53  & 0.90 \\
                            \cline{2-5}
\multirow{ 3}{*}{\textbf{Proposed- Shared}} & Step $t$ & 0.62   & 0.57  & 0.85\\
                                    & Step $t-1$ & 0.61   & 0.56  & 0.85  \\
                                    & Step $t-2$ & 0.57  & 0.52 & 0.88 \\
                            \cline{2-5}
\multirow{ 3}{*}{\textbf{Pre-LSTM}} & Step $t$ & 0.48   & 0.43  & 0.90\\
& Step $t-1$          & 0.48   & 0.43  & 0.90  \\
& Step $t-2$           &  0.43  & 0.43  & 0.94 \\
\hline
\end{tabular}
\end{table}

\begin{table}[H]
\centering
\caption{Performance of our \textbf{proposed model} architecture for \textbf{prediction} at each timestep output by the LSTM}
\label{table:prediction_timestep}
\begin{tabular}{@{}llccc@{}}
\toprule
                                                   &          & \textbf{F1 Score} & \textbf{IOU Score} & \textbf{Loss (Focal + Dice)} \\ \midrule
\multirow{3}{*}{\textbf{Prediction Task-Unshared}} & Step $t-1$ & 0.56              & 0.51               & 0.90                         \\
                                                   & Step $t-2$ & 0.56              & 0.52               & 0.90                         \\
                                                   & Step $t-3$ & 0.50              & 0.47               & 0.90                         \\ \cmidrule(l){2-5} 
\multirow{3}{*}{\textbf{Prediction Task-Shared}}            & Step $t-1$ & 0.57              & 0.53               & 0.89                         \\
                                                   & Step $t-2$ & 0.56              & 0.52               & 0.89                         \\
                                                   & Step $t-3$ & 0.53              & 0.50               & 0.89                         \\ \midrule
\multirow{3}{*}{\textbf{Prediction Task-Unshared}} & Step $t-2$ & 0.47              & 0.44               & 0.93                         \\
                                                   & Step $t-3$ & 0.48              & 0.45               & 0.93                         \\
                                                   & Step $t-4$ & 0.45              & 0.42               & 0.93                         \\ \cmidrule(l){2-5} 
\multirow{3}{*}{\textbf{Prediction Task-Shared}}   & Step $t-2$ & 0.45              & 0.42               & 0.93                         \\
                                                   & Step $t-3$ & 0.48              & 0.46               & 0.93                         \\
                                                   & Step $t-4$ & 0.44              & 0.41               & 0.93                         \\ \bottomrule
\end{tabular}
\end{table}

\newpage
\section{Supplemental model figures}
Alternative model architectures explored during this study.

\begin{figure}[H]
    \centering
    \includegraphics[width=0.8\linewidth, trim={0 0cm 0 0cm}]{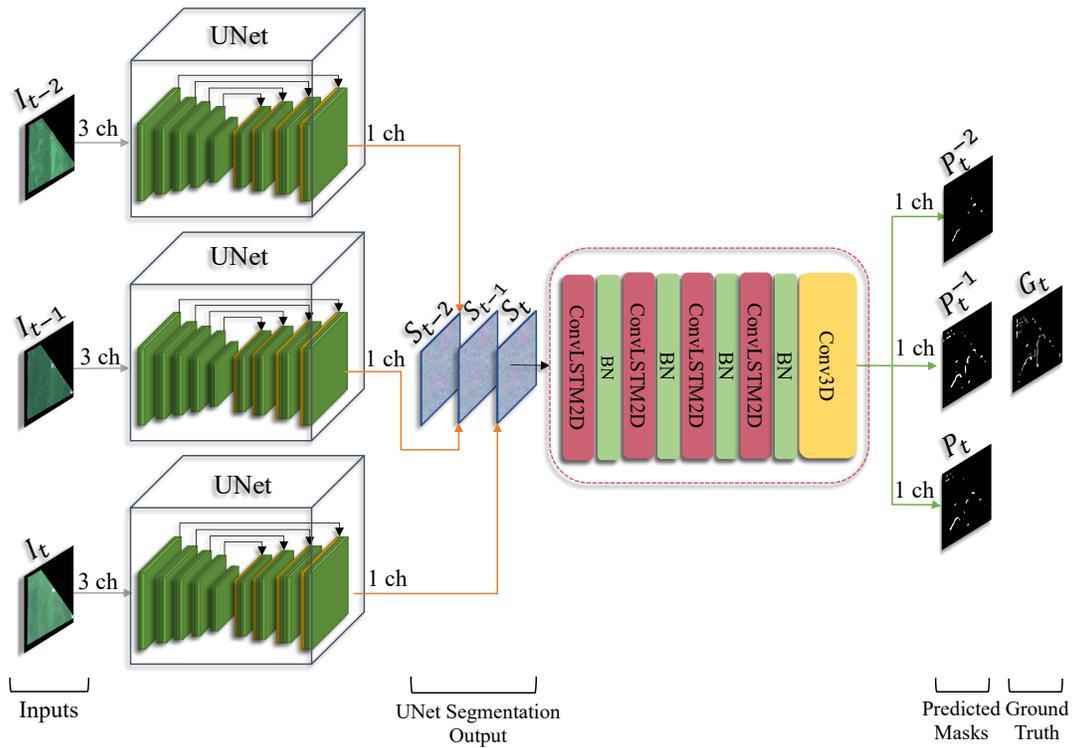}
    \caption{Diagram of our \textit{Proposed (Longitudinal) Model} for reference.  Same diagram as Figure 3 in the main text. [F1: 0.62, IOU: 0.57].}  
    
\label{fig:post_lstm}
\end{figure}

\begin{figure}[hbt]
    \centering
    \includegraphics[width=0.6\linewidth, trim={0 0cm 0 0cm}]{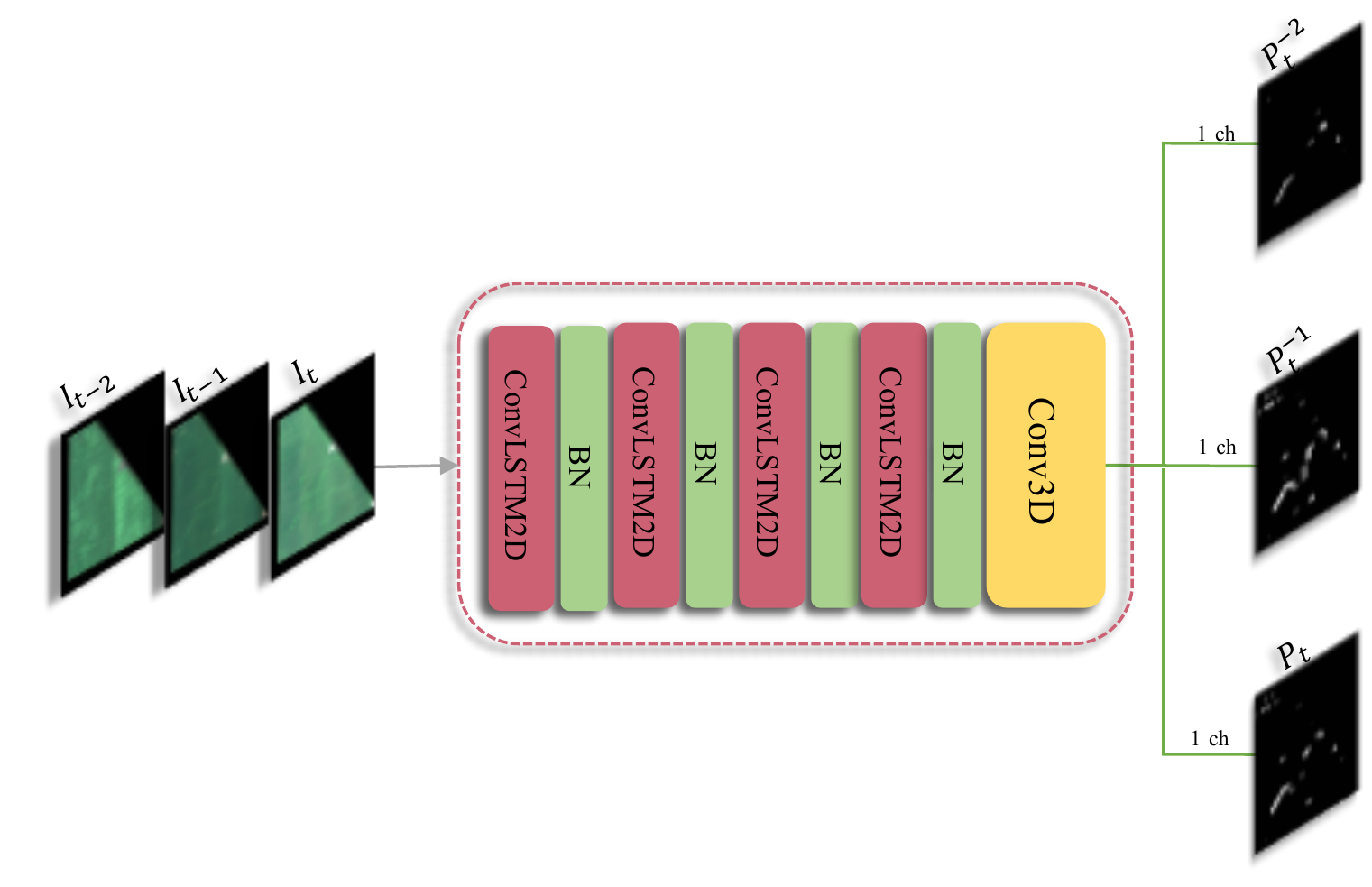}
    \caption{Diagram of our \textit{Only-LSTM} model which passes the three input images directly through the convolutional LSTM. [F1: 0.48, IOU: 0.43]}   
    
\label{fig:only_lstm}
\end{figure}

\begin{figure}[hbt!]
    \centering
    \includegraphics[width=0.6\linewidth, trim={0 0cm 0 0cm}]{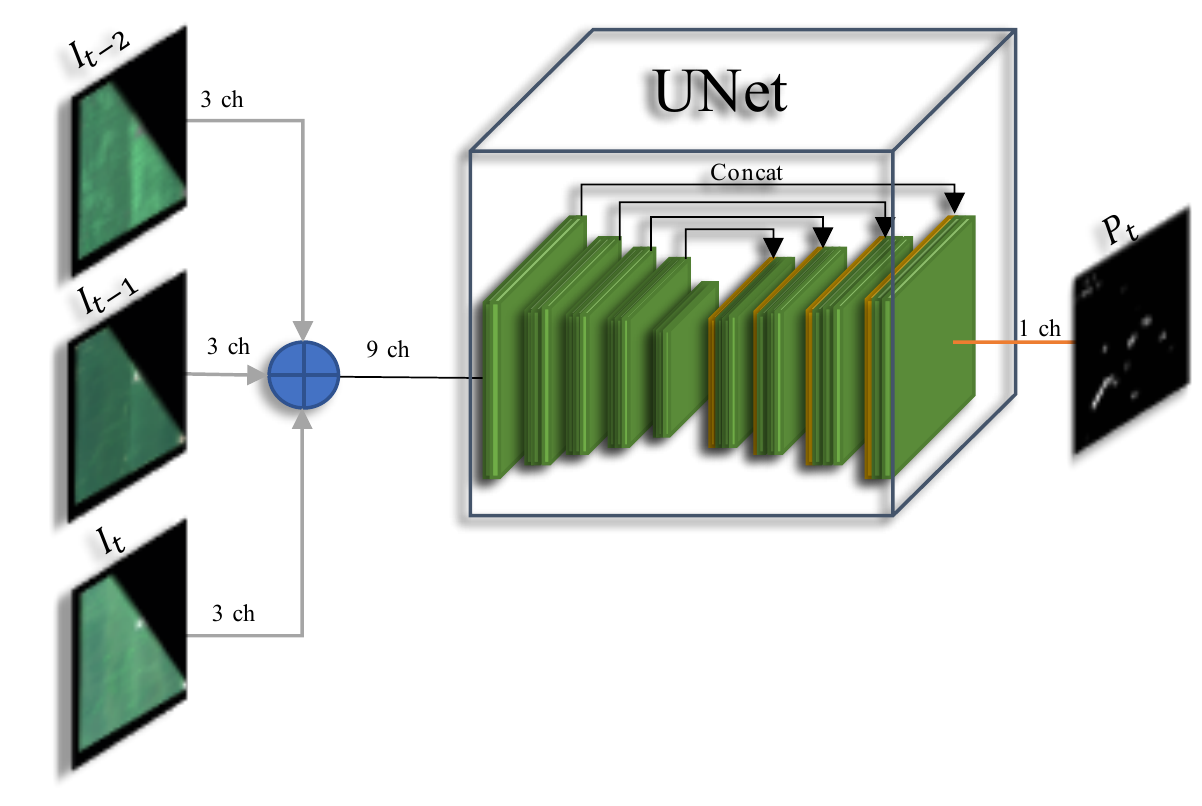}
    \caption{Diagram of our \textit{9-Channel} model which concatenates the input images to a 9-channel input before passing through the UNet. [F1: 0.23, IOU: 0.15]}   
    
\label{fig:9_channel}
\end{figure}

\begin{figure}[hbt!]
    \centering
    \includegraphics[width=0.6 \linewidth, trim={0 0cm 0 0cm}]{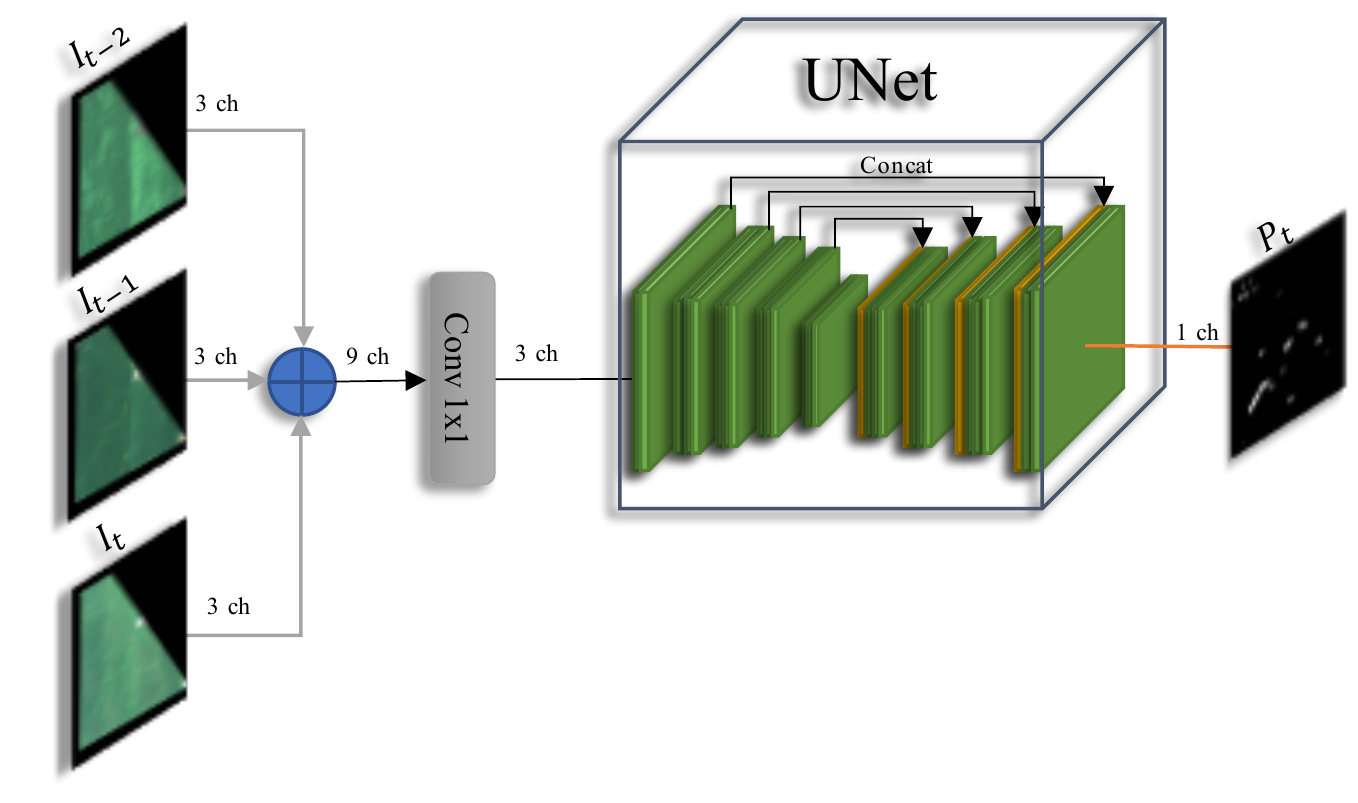}
    \caption{Diagram of our ~\textit{9-Channel + 1D Conv} model which concatenates the input images to a 9-channel and then applies a 1D convolution to reduce the dimensionality before passing through the UNet. [F1: 0.48, IOU: 0.34]}   
    
\label{fig:9_channel_1D_conv}
\end{figure}

\begin{figure}[hbt!]
    \centering
    \includegraphics[width=0.6\linewidth, trim={0 0cm 0 0cm}]{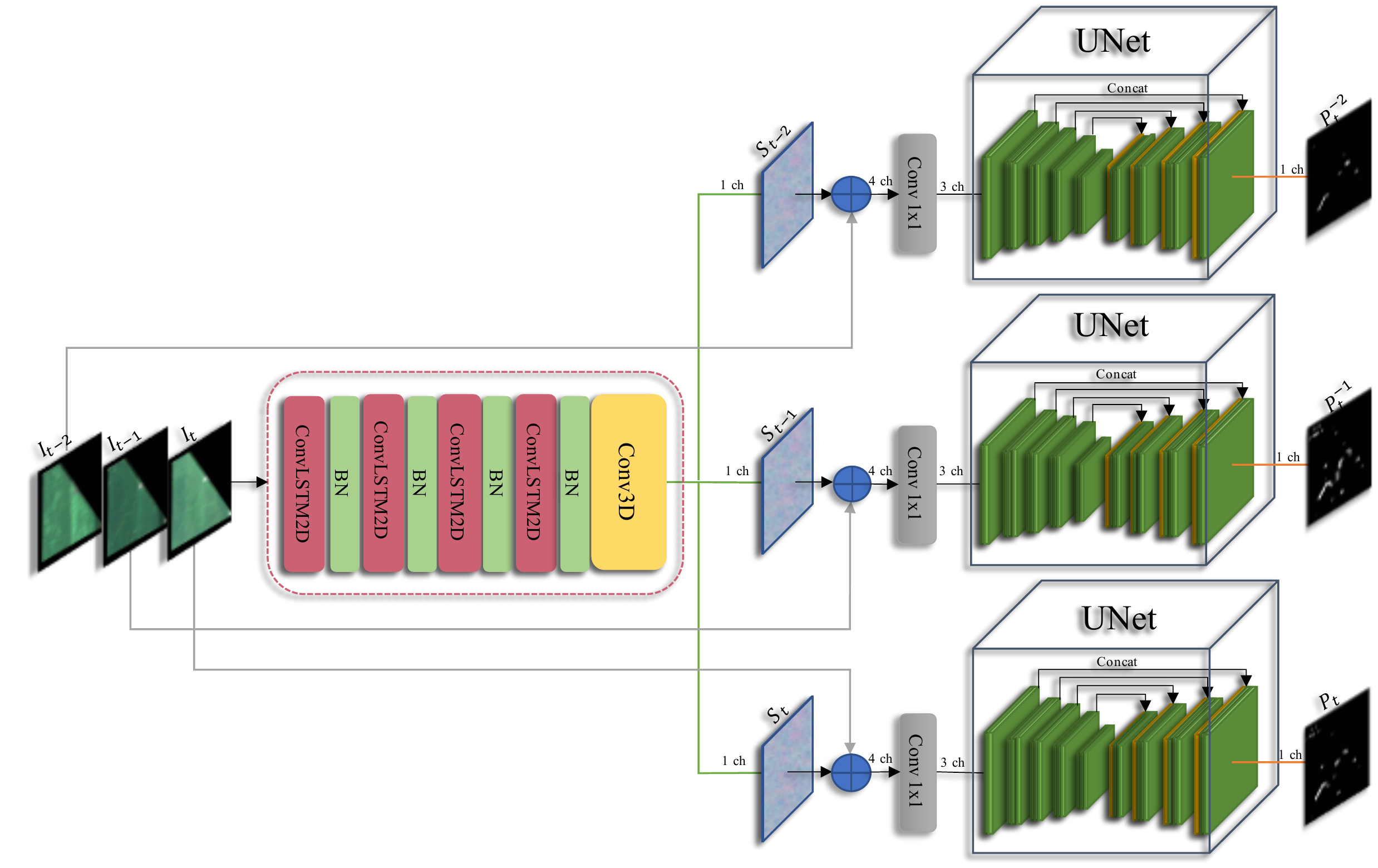}
    \caption{Our ~\textit{Pre-LSTM + Concatenation} architecture which passes the images through the convolutional LSTM to generate an intermediate mask $S_{t-n}$ for each.  That mask is then appended to the corresponding image ($I_{t-n}$) via concatenation, and those 4-channels passed through a UNet to generate a final prediction from each image. [F1: 0.42, IOU: 0.29]}   
    
\label{fig:pre_lstm_concat}
\end{figure}
\begin{figure}[hbt!]
    \centering
    \includegraphics[width=0.6\linewidth, trim={0 0cm 0 0cm}]{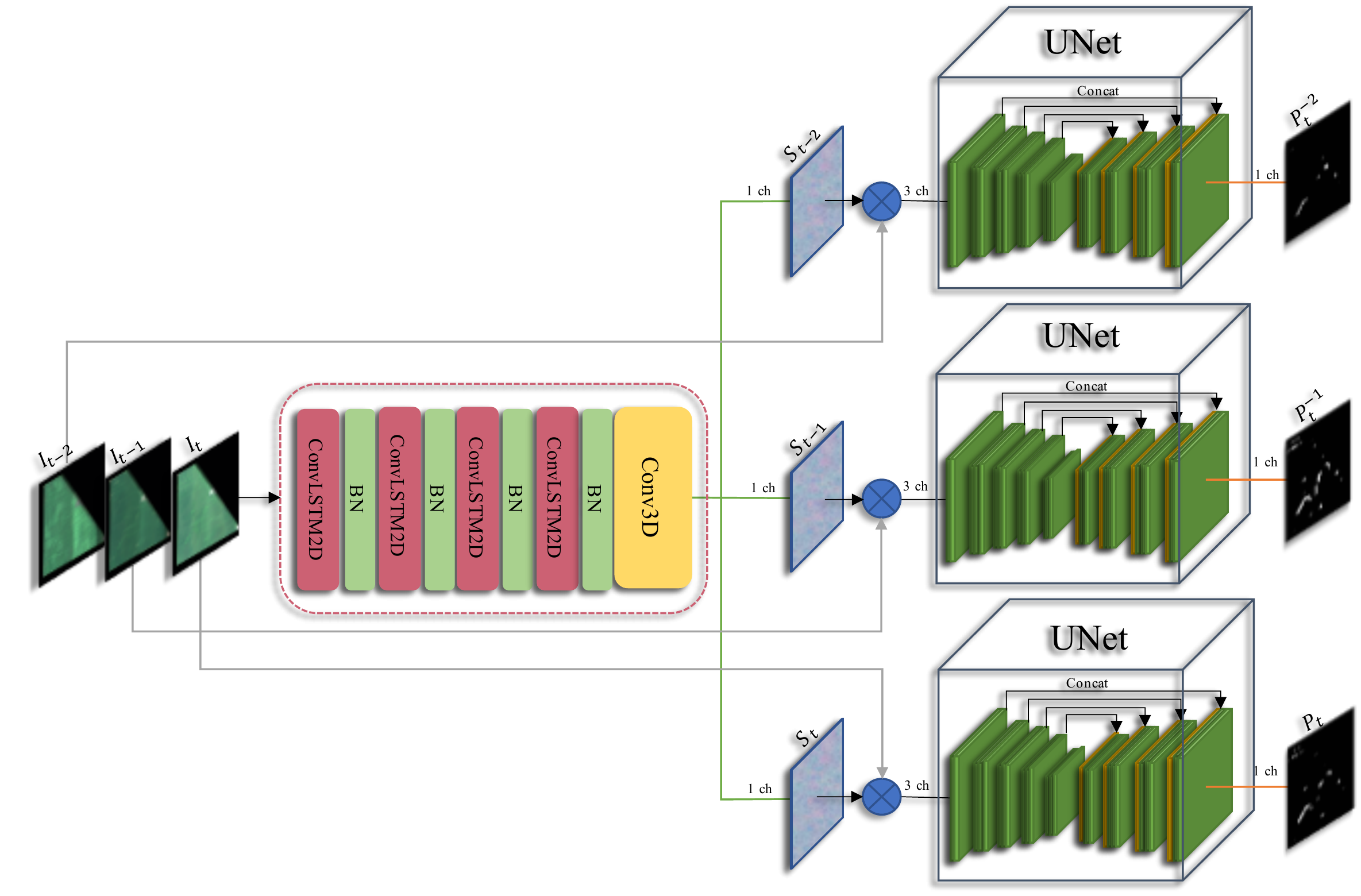}
    \caption{Our ~\textit{Pre-LSTM + Multiplication} architecture is similar to the above, but combines the intermediate prediction $S_{t-n}$ with the image $I_{t-n}$ through the Hadamard product. [F1: 0.42, IOU: 0.29]}   
    
\label{fig:pre_lstm_multi}
\end{figure}

\begin{figure}[hbt!]
    \centering
    \includegraphics[width=0.4\linewidth, trim={0 0cm 0 0cm}]{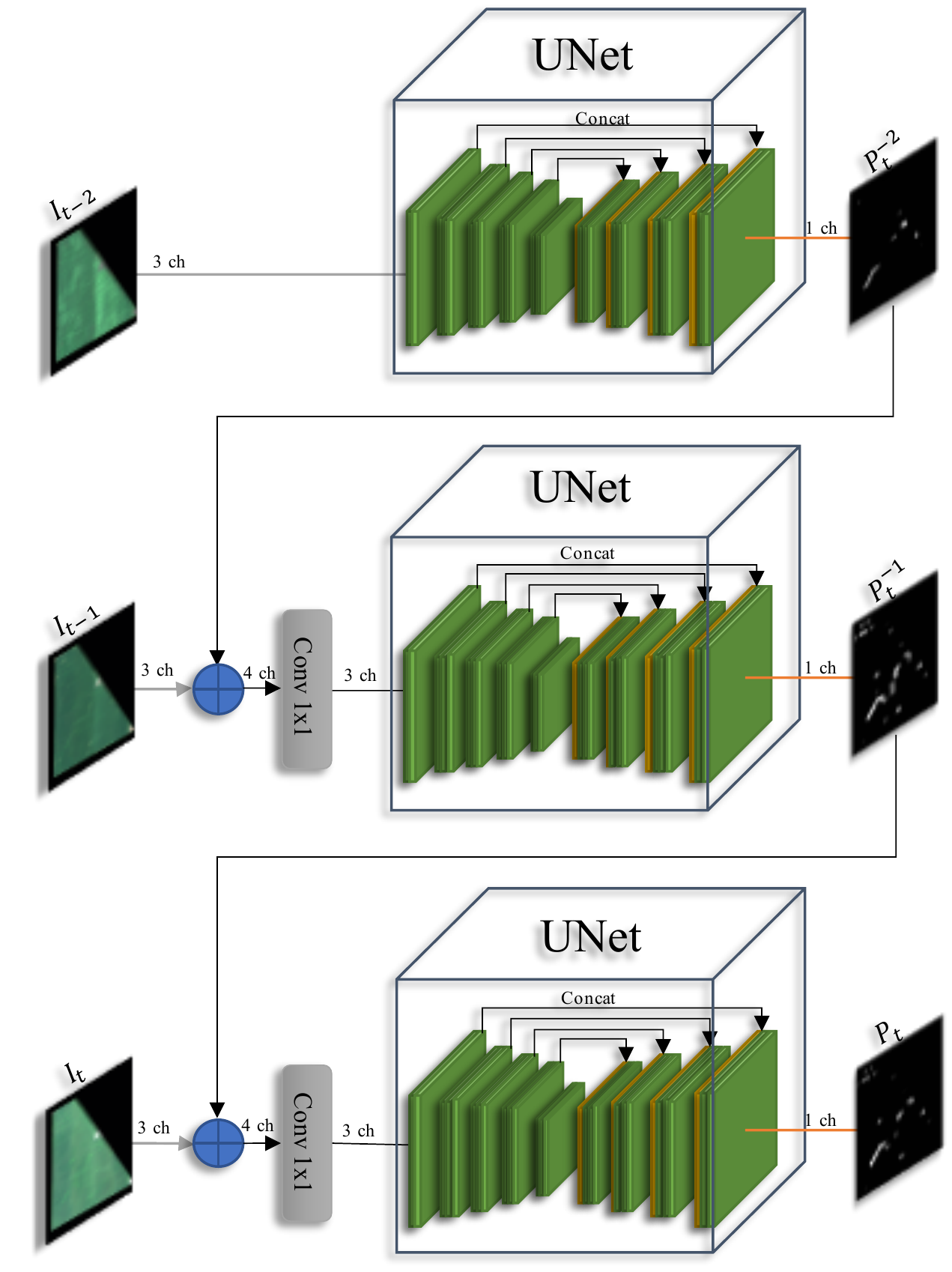}
    \caption{Diagram of our \textit{Cascading-Model + Concatenation} architecture which passes the first image through the UNet to generate a prediction.  This prediction is combined with the second image via concatenation and that 4-channel input is passed through a second UNet to generate an updated prediction.  That prediction is combined with the final image, passed through the UNet, and the final output prediction is returned.  The total loss is the combined focal + dice loss for all three predicted masks. [F1: 0.43, IOU: 0.30]}   
    
\label{fig:cascading_concat}
\end{figure}

\begin{figure}[hbt!]
    \centering
    \includegraphics[width=0.4\linewidth, trim={0 0cm 0 0cm}]{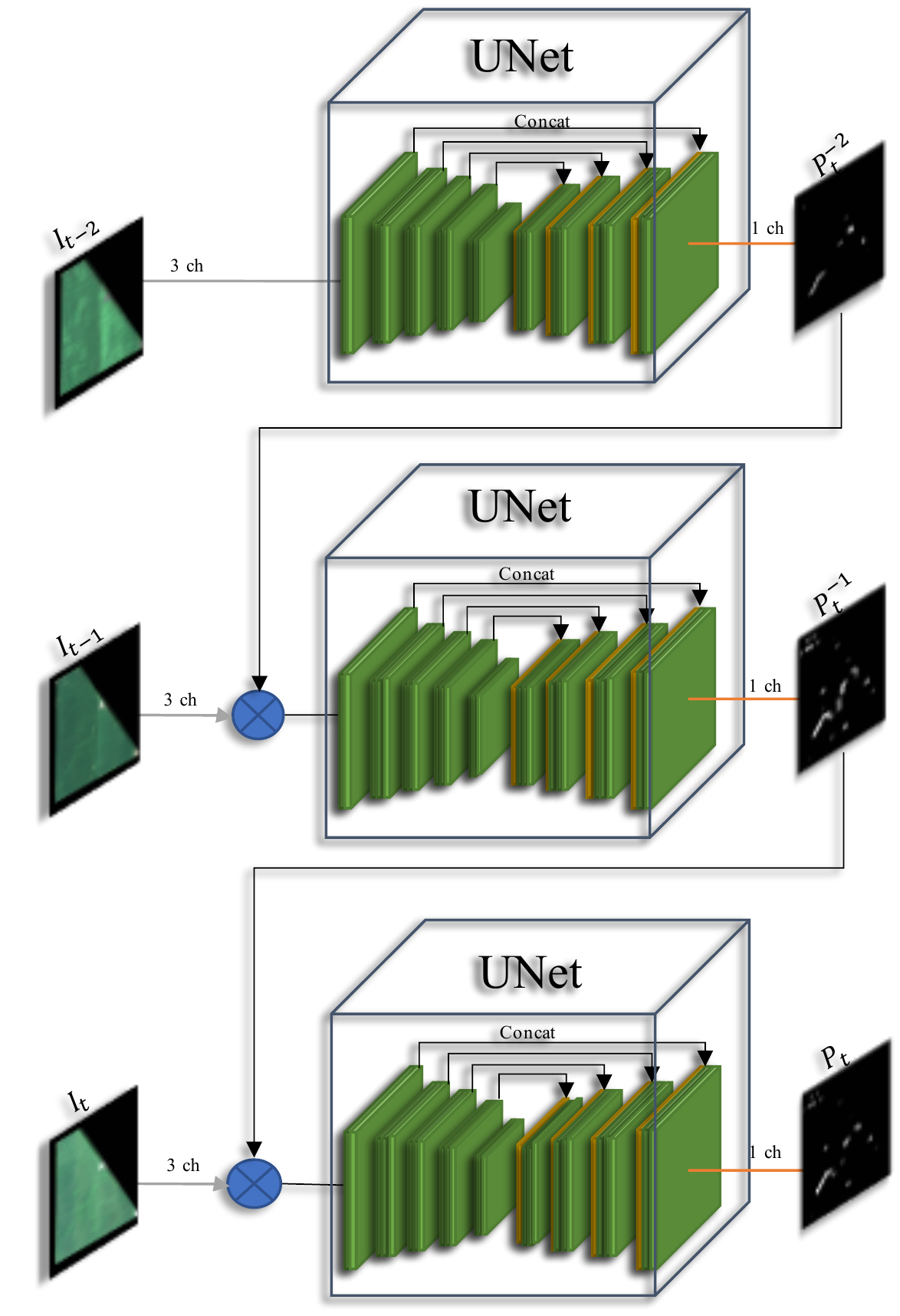}
    \caption{Diagram of our \textit{Cascading-Model + Multiplication} architecture which functions similarly to the model above, but combines the intermediary masks with the following image via the Hadamard product as opposed to concatenation. [F1: 0.38, IOU: 0.25]}   
    
\label{fig:cascading_multi}
\end{figure}

\clearpage
\section{Supplemental result figures}

\begin{figure}[H]
    \centering
    \includegraphics[width=0.9\linewidth, trim={0 0cm 0 0cm}]{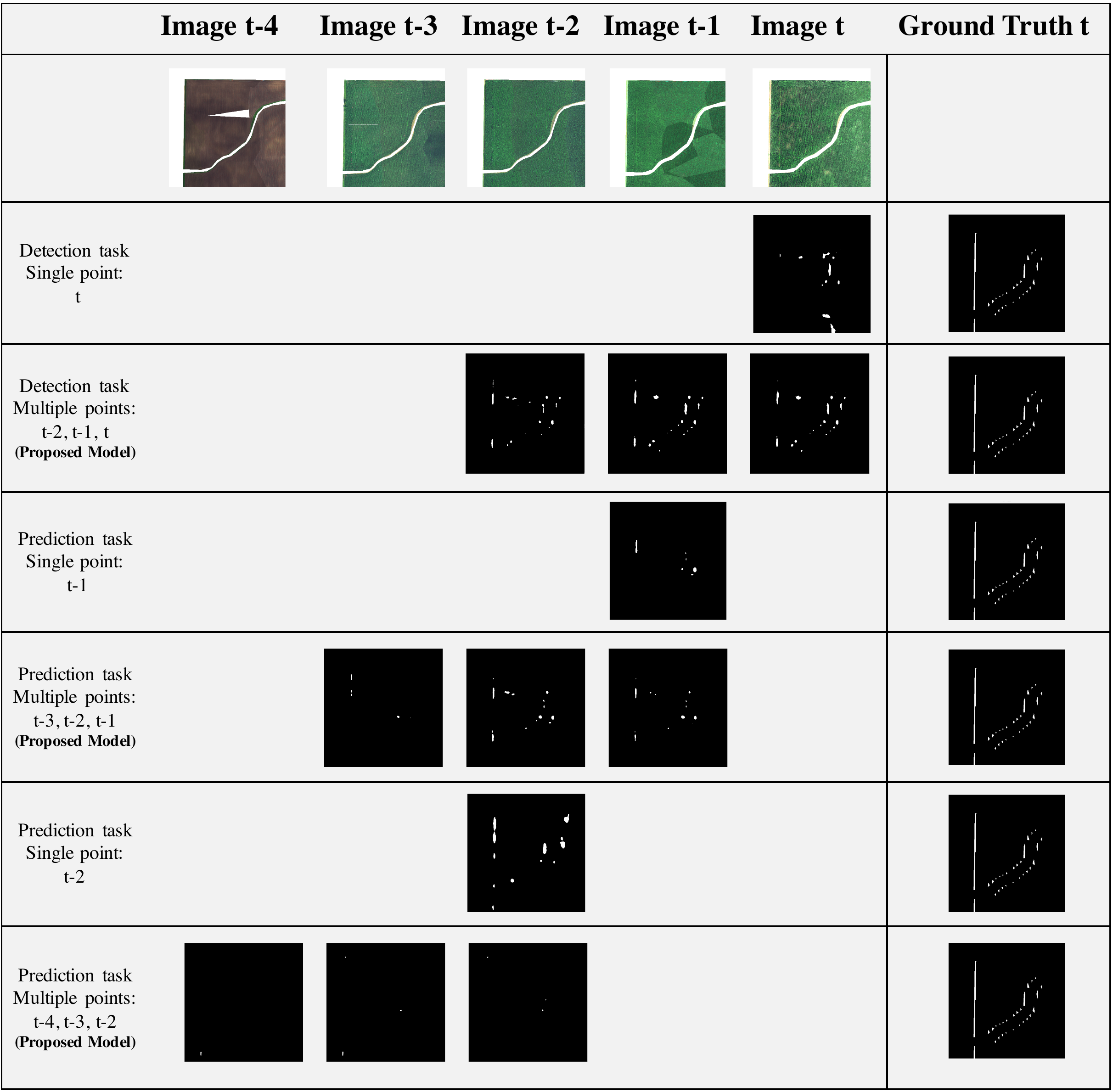}
    \caption{Additional results analogous to Figure 4 in the main text but for a different field.  The detection model continued to perform well although the prediction models struggled on this example.}   
    
\label{fig:results_2}
\end{figure}